\documentclass[11pt]{article}

\usepackage[preprint]{acl}

\usepackage{times}
\usepackage{latexsym}
\usepackage{booktabs}
\usepackage{url}
\usepackage{amssymb}
\usepackage[T1]{fontenc}

\usepackage[utf8]{inputenc}

\usepackage{microtype}

\usepackage{inconsolata}

\usepackage{graphicx}
\usepackage{pifont}
\newcommand{\cmark}{\ding{51}}
\newcommand{\xmark}{\ding{55}}

\title{SemEval-2026 Task 12: Abductive Event Reasoning: Towards Real-World Event Causal Inference for Large Language Models}

\author{
\textbf{Pengfei Cao\textsuperscript{1,2}}\thanks{Equal contribution.}, 
\textbf{Mingxuan Yang\textsuperscript{1,2}}\footnotemark[1],
\textbf{Yubo Chen\textsuperscript{1,2}},  \\
\textbf{Chenlong Zhang\textsuperscript{1,2}}, 
\textbf{Mingxuan Liu\textsuperscript{1,2}}, 
\textbf{Kang Liu\textsuperscript{1,2}}, 
\textbf{Jun Zhao\textsuperscript{1,2}}\thanks{Corresponding author.} \\
\textsuperscript{1}The Key Laboratory of Cognition and Decision Intelligence for Complex Systems,\\
Institute of Automation, Chinese Academy of Sciences, Beijing, China\\
\textsuperscript{2}School of Artificial Intelligence, University of Chinese Academy of Sciences, Beijing, China\\
\small{\{yangmingxuan2025, zhangchenlong2024, liumingxuan2025\}@ia.ac.cn} \\
\small{\{pengfei.cao, yubo.chen, kliu, jzhao\}@nlpr.ia.ac.cn}
}

\begin{document}
\maketitle

\begin{abstract}
Understanding why real-world events occur is important for both natural language processing and practical decision-making, yet direct-cause inference remains underexplored in evidence-rich settings. To address this gap, we organized SemEval-2026 Task 12: Abductive Event Reasoning (AER).\footnote{The task data is available at \url{https://github.com/sooo66/semeval2026-task12-dataset.git}} The task asks systems to identify the most plausible direct cause of a target event from supporting evidence. We formulate AER as an evidence-grounded multiple-choice benchmark that captures key challenges of real-world causal reasoning, including distributed evidence, indirect background factors, and semantically related but non-causal distractors. The shared task attracted 122 participants and received 518 submissions. This paper presents the task formulation, dataset construction pipeline, evaluation setup, and system results. AER provides a focused benchmark for abductive reasoning over real-world events and highlights challenges for future work on causal reasoning and multi-document understanding.
\end{abstract}

\section{Introduction}


Understanding why real-world events occur is important for both natural language processing and practical decision-making \citep{waldmann2013causal}. Identifying direct causes, however, remains difficult in realistic settings because evidence is often incomplete, distributed across multiple documents, and mixed with background conditions, consequences, or semantically related but non-causal events \citep{tu-etal-2019-multi,min-etal-2025-multi,yang-etal-2018-hotpotqa,trivedi-etal-2022-musique}. Despite progress in event causality identification \citep{cao2021knowledge, zuo2021improving, he2024zero, mirza-2014-extracting,romanou-etal-2023-crab}, event relation extraction \citep{cao2021uncertainty, hao2023complex, sui2023joint, tang-etal-2021-discourse,wang-etal-2022-maven}, and abductive reasoning \citep{bhagavatula2020abductive,ponti-etal-2020-xcopa}, direct-cause inference under noisy multi-document evidence remains underexplored.

\begin{figure}[t]
\centering
\includegraphics[width=\linewidth]{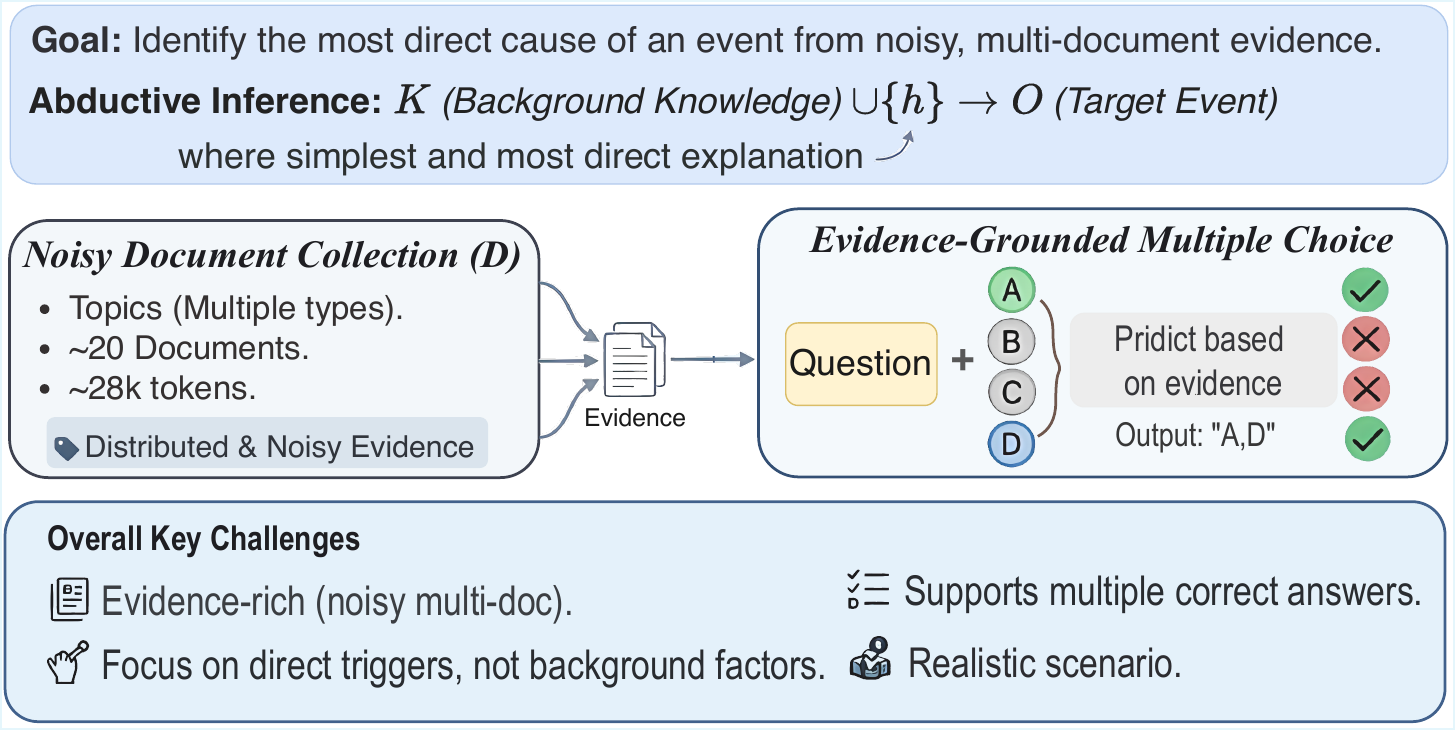}
\caption{Overview of the Abductive Event Reasoning (AER) task. Given a noisy multi-document evidence collection, the goal is to identify the most plausible direct cause of a target event via evidence-grounded abductive inference. The task is challenging because the supporting evidence is distributed and noisy, systems must focus on direct triggers rather than background conditions, and multiple answer options may be correct.}
\label{fig:ow}
\end{figure}


To address this gap, we organized SemEval-2026 Task 12: Abductive Event Reasoning (AER), a shared task that asks systems to identify the most plausible direct cause of a target event from supporting evidence (Figure~\ref{fig:ow}). We formulate AER as an evidence-grounded multiple-choice benchmark for real-world event causal reasoning. Unlike traditional causal relation extraction, which often focuses on predefined or locally expressed event pairs, AER requires systems to reason over noisy multi-document evidence, distinguish direct causes from broader contextual factors, and avoid semantically plausible but causally incorrect distractors \citep{romanou-etal-2023-crab,trivedi-etal-2022-musique}.

To support this task, we constructed a benchmark through a multi-stage pipeline involving document collection, event extraction, timeline construction, multi-model causality scoring, and human verification. The resulting dataset contains 60 topics and 2,831 question instances derived from real-world reports spanning 2016 to 2025. On average, each topic contains 19.7 documents, with supporting evidence averaging approximately 28K tokens. The difficulty of the task arises not only from causal ambiguity but also from the scale and noisiness of the evidence, which place substantial demands on long-context understanding and cross-document integration \citep{bai-etal-2024-longbench,trivedi-etal-2022-musique}.

The shared task attracted 122 participants and received a total of 518 submissions. Ultimately, 21 teams submitted system description papers. This paper presents the overall task design, dataset construction details, and key characteristics of the participating systems. The submitted systems demonstrate how current machine learning techniques perform when addressing complex causal reasoning under long, noisy multi-document evidence, and they provide valuable insights for future improvements in causal reasoning, multi-document understanding, and robust abductive reasoning.

\section{Related Work}

\paragraph{Event Causality Reasoning.}
A substantial body of prior work has studied causality in text, including temporal and causal relation extraction \citep{mirza-tonelli-2014-analysis}, unified event relation extraction \citep{wang-etal-2022-maven}, and the construction of causal reasoning benchmarks \citep{romanou-etal-2023-crab}. Much of this literature focuses on relatively local settings, such as sentence-level or document-level relation classification \citep{mirza-tonelli-2014-analysis,zhao-etal-2021-graph,liu-etal-2024-identifying}, often assuming explicit relation instances or predefined event pairs. More recent work has moved toward richer causal benchmarks that consider contextualized event relations and graded causal strength \citep{romanou-etal-2023-crab,yuan-etal-2023-csb}. In particular, recent studies highlight that causal judgments over real-world events are often context-dependent, temporally constrained, and more complex than simple binary cause--effect detection \citep{halpern2016actual,pearl2009causality}. Our task is closely related to this line of work but differs in formulation: rather than classifying predefined event pairs, AER asks systems to select the most plausible direct cause of a target event from multiple evidence-grounded candidates.

\paragraph{Multi-document Reasoning.}
Our task is also related to research on reasoning over information distributed across multiple documents. Prior work has shown that language understanding becomes significantly more difficult when evidence must be aggregated from multiple sources rather than read from a single passage \citep{tu-etal-2019-multi,yang-etal-2018-hotpotqa}. This challenge is particularly relevant for causal reasoning over real-world events, where supporting evidence may be scattered across reports published at different times and mixed with noise or irrelevant background information. Although recent work has explored cross-document or real-world causal reasoning benchmarks \citep{romanou-etal-2023-crab,min-etal-2025-multi}, these datasets are often smaller or more structured. Related work on multi-hop QA and long-context evaluation has likewise shown that models remain vulnerable to shortcut reasoning and degradation over long, compositional evidence chains \citep{trivedi-etal-2022-musique,bai-etal-2024-longbench}. In contrast, AER introduces a larger and more realistic noisy multi-document environment, significantly increasing the difficulty of long-context comprehension, noise robustness, and fine-grained causal discrimination. This makes AER a more challenging and representative benchmark for real-world event causal reasoning.

\section{Task Description}

\subsection{Task Overview}

SemEval-2026 Task 12: {Abductive Event Reasoning (AER)} aims to evaluate whether systems can identify the most plausible and most direct cause of a target event from supporting evidence. The task is grounded in the concept of \textit{abductive reasoning}, which involves finding the best explanation for an observed outcome \citep{walton2014abductive}. Formally, let $O$ denote the observed event, $K$ denote background knowledge, and $H$ denote the set of candidate hypotheses derived from the available context. The goal of abductive reasoning is to select the simplest and most direct hypothesis $h \in H$ such that

\[
K \cup \{h\} \models O
\]

That is, the hypothesis together with the background knowledge forms an explanation for the observed outcome. In our setting, the \textit{target event} corresponds to the observed outcome, the \textit{candidate causes} correspond to explanatory hypotheses supported by the provided evidence, and the model's internal knowledge and reasoning capability correspond to the background knowledge.

A central design principle of AER is the distinction between \textbf{direct causes} and broader background factors. Real-world events are often associated with many temporally related or semantically similar developments, but not all of them provide explanations of equal strength \citep{halpern2016actual}. For example, when the target event is “severe flooding in a certain region,” the evidence may include both “prolonged heavy rainfall” and “global climate warming trends.” The task requires the system to prioritize identifying “heavy rainfall” as the immediate triggering factor rather than treating climate warming as an equally direct cause, even though it may be relevant in a broader context. In short, AER focuses on causes that \textbf{directly trigger} the target event, rather than factors that merely provide background conditions. This distinguishes the task from simple topic matching and also makes it more challenging than traditional sentence-level causal relation extraction.

\subsection{Task Formalization}

We formulate AER as an \textbf{evidence-based multiple-choice task} that allows \textbf{multiple correct answers}. Each instance is defined as a triple $I = (e_t, D, C)$, where $e_t$ is the target event, $D = \{d_1, d_2, \ldots, d_n\}$ is an unordered set of supporting documents, and $C = \{c_A, c_B, c_C, c_D\}$ is a set containing exactly four candidate options. The system is required to output a subset of option labels indicating all correct direct causes.

Allowing multiple correct answers is an intentional design choice rather than an annotation artifact. Because causality is inherently non-binary, a target event in real-world event chains may have more than one directly relevant triggering factor that is clearly supported by the evidence \citep{halpern2016actual,romanou-etal-2023-crab}. Therefore, this benchmark evaluates whether systems can recover a \textit{complete and plausible set of direct causes}, rather than forcing the problem into a single-answer format.

The official shared task consists of a single track. Participants are provided with the \textit{full document collections} rather than curated timelines or summarized evidence. On average, each topic contains 19.7 documents, with an average topic-level evidence length of 28,047 tokens. Importantly, these document collections are \textbf{intentionally noisy}: besides relevant reports, they may also include documents that are topically similar but causally irrelevant. The inclusion of such distractors is designed to better approximate real-world evidence conditions and to reduce the effectiveness of shallow lexical matching strategies.

\subsection{Input and Output Format}

Each problem instance contains exactly four candidate options labeled \texttt{A} to \texttt{D}. System outputs should be a comma-separated string of option labels, for example \texttt{B,C}. Since multiple answers may be correct, predictions may include one or more labels. Predictions must only contain valid labels from \texttt{A--D}; invalid outputs receive a score of 0 for the corresponding instance.

For each instance, the evidence is provided as a collection of documents associated with the corresponding topic. Documents are treated as an unordered set, and participating systems must determine which options are directly supported as causes of the target event based on these documents.

\subsection{Evaluation Metrics}

The official evaluation metric is \textbf{instance-based accuracy}, which accounts for both exact matches and partial matches. Let $G$ denote the gold set of correct options and $P$ denote the system prediction set. The score for a single instance is defined as follows:

\begin{itemize}
\item \textbf{1.0 (Exact Match)}: if $P = G$;
\item \textbf{0.5 (Partial Match)}: if $P$ is a non-empty proper subset of $G$;
\item \textbf{0.0 (Incorrect)}: all other cases, including when $P$ contains any incorrect option or when $P$ is empty.
\end{itemize}

The final system score is the average of the instance-level scores across all evaluation instances. This scoring scheme encourages precise causal predictions and penalizes over-prediction, which is particularly important in AER since some instances contain multiple correct answers.

\subsection{Task Example}

Table~\ref{tab:aer-example} presents an example from the benchmark. The target event corresponds to an observed outcome, the evidence is provided by a set of retrieved documents, and the candidate options include both plausible and implausible explanations. In this example, the direct cause is the shooting incident, while the other options correspond to indirect factors, subsequent events, or unrelated distractors.

\begin{table*}[t]
\centering
\small
\setlength{\tabcolsep}{6pt}
\renewcommand{\arraystretch}{1.2}
\begin{tabular}{p{0.14\textwidth} p{0.80\textwidth}}
\toprule
\textbf{Target Event} & Videos of the assassination circulated on social media. \\
\midrule
\textbf{Context} & A collection of documents containing mixed relevant and irrelevant evidence \\
\midrule
\multicolumn{2}{l}{\textbf{Options}} \\
\midrule
A \xmark & The shooter used a handmade gun. \\
B \xmark & Security arrested the suspected gunman, Tetsuya Yamagami. \\
C \xmark & Shinzo Abe became the deputy chief cabinet secretary in the early 2000s. \\
D \cmark & A man fired twice at Shinzo Abe. \\
\midrule
\textbf{Answer} & \textbf{D} \\
\bottomrule
\end{tabular}
\caption{An illustrative example from the AER dataset. The target event concerns the circulation of assassination videos on social media, and the candidate causes range from peripheral background information to the most immediate precipitating event. This example highlights the task’s focus on selecting the most plausible direct cause rather than merely related context.}
\label{tab:aer-example}
\end{table*}

\section{Dataset Construction}

The AER benchmark is derived from real-world news reports published between 2016 and 2025. The dataset was constructed through a multi-stage pipeline that combines large language models with human verification, with the goal of producing evidence-based multiple-choice questions for direct cause inference.

\subsection{Overall Pipeline}

\begin{figure*}[t]
\centering
\includegraphics[width=1.0\textwidth]{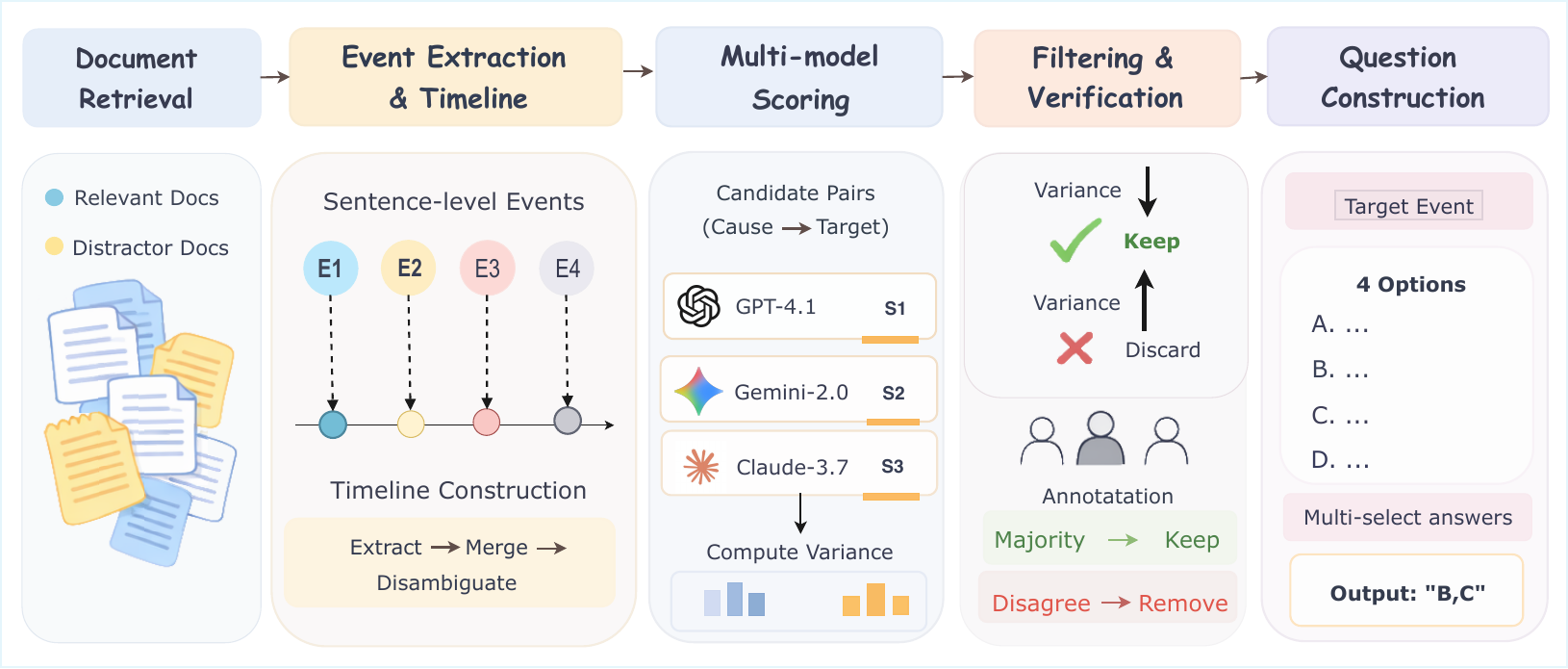}
\caption{Construction pipeline of the AER benchmark.}
\label{fig:pipeline}
\end{figure*}

Inspired by the framework of \citet{romanou-etal-2023-crab}, we designed the data construction pipeline shown in Figure~\ref{fig:pipeline}. We first collected topic-centered document sets using the Google News API. We then extracted sentence-level events from the retrieved documents and organized them into topic-level timelines. Next, we scored candidate causal relations using multiple language models and combined these scores with human verification. Finally, the verified event pairs were converted into multiple-choice question instances.

\subsection{Data Collection and Event Extraction}

Our source documents were collected through the Google News API.\footnote{\url{https://serper.dev/}} We retrieved news reports centered on specific topics and filtered the results using fixed time windows. The resulting document sets were intended to cover both the core trajectory of each event and its surrounding background context. To make the task more realistic and reduce the effectiveness of shallow lexical matching strategies, we deliberately introduced distractor documents. These documents were obtained by querying with keywords that were semantically related to the topic but tended to lead to reports outside the target event chain.

For the retrieved documents, we used GPT-4.1 \citep{openai2025gpt41} to extract sentence-level event mentions. The extracted outputs were then manually inspected to remove malformed or low-quality results. In total, the initial collection contained 913 unique event mentions, which were subsequently normalized into 865 timeline-level events. Sentence-level event extraction provides a practical balance between granularity and scalability: it is fine-grained enough to support causal reasoning, while keeping the complexity of subsequent alignment and annotation stages manageable.

\subsection{Timeline Construction}

After event extraction, we used the then state-of-the-art large language model GPT-4.5 \citep{openai2025gpt45} to organize event mentions into topic-level timelines, followed by human verification. During this process, mentions that referred to the same real-world event but were expressed differently were merged and disambiguated through a combination of prompt-based LLM processing and manual validation.

\subsection{Causality Scoring and Human Verification}

For each target event, we treated all events occurring earlier on the timeline as candidate causes. Each candidate event pair was then scored by three models: GPT-4.1 \citep{openai2025gpt41}, Gemini-2.0-Flash \citep{google2025gemini20flash}, and Claude-3.7-Sonnet \citep{anthropic2025claude37sonnet}. Each model produced a causality score between 0 and 100. We then computed the variance across the three model scores for each candidate pair and used this signal to identify uncertain cases prior to human annotation.

The remaining candidate pairs were manually verified by three annotators. Each event pair was independently labeled by all three annotators. Samples with complete disagreement among the three annotators were discarded, whereas samples with a majority decision were retained. In the final release, we use Krippendorff's $\alpha$ as the primary inter-annotator agreement metric \citep{krippendorff2011computing}. Final agreement statistics are reported in Section~\ref{sec:annotation_quality}.

\subsection{Question Generation}

Strictly verified causal event pairs were used to construct the final multiple-choice questions. To ensure task difficulty and prevent shortcut learning, we adopted a difficulty-aware stratified distractor design strategy. Specifically, for each target event $E_t$, we constructed a positive candidate pool ($P$) and a negative distractor pool ($N$) based on the timeline and causal labels.

The distractor pool $N$ was further divided into three categories: (1) \textbf{temporal distractors}: events occurring after $E_t$ on the timeline, i.e., outcomes or subsequent developments; (2) \textbf{semantic distractors}: events that share salient entities or keywords with $E_t$ but do not bear a substantive causal relation in the given context; and (3) \textbf{indirect/background factors}: events that received relatively low scores during the initial scoring stage and typically correspond to long-term background conditions or indirect causes.

During instance assembly, we sampled options to create four-choice (A/B/C/D) test items for each $E_t$. In addition, we introduced \textit{None of the others are correct causes.} instances to test model abstention ability and mitigate hallucination.

\subsection{Dataset Statistics}

The final benchmark contains 60 topics and 2,831 question instances. The training, development, and test sets contain 1,819, 400, and 612 instances, respectively. The benchmark is built on topic-organized document collections spanning real-world reports from June 2016 to November 2025, and includes 913 unique event mentions and 865 timeline events.

On average, each topic contains 19.7 documents, and the average topic-level evidence length is 28,047 tokens; the longest topic reaches 91,417 tokens. The average document length is 1,088.6 tokens. These statistics indicate that AER not only covers a wide range of topics, but also involves lengthy and highly variable contextual evidence, posing a substantial challenge for long-context understanding.

AER is designed as a genuinely multi-answer benchmark rather than a traditional single-label multiple-choice task. In the training and development sets, 56.42\% of instances have only one correct answer, whereas 43.58\% contain multiple correct answers; the average number of gold labels per instance is 1.57. This shows that multi-answer reasoning is not a marginal phenomenon, but a core property of the benchmark.

Regarding option distribution, after aggregating all positive labels, the positions of correct answers across A--D are relatively balanced, indicating that models cannot benefit from fixed positional biases. At the topic level, representative topic timestamps span multiple years, showing that the benchmark is not concentrated within a narrow time range, but is instead grounded in news collections reflecting long-term real-world developments.

The document length distribution further shows that although most documents remain within a manageable range, a substantial proportion of long documents still exists, increasing the difficulty of evidence retrieval and direct cause identification.

\begin{table}[t]
\centering
\small
\begin{tabular}{lc}
\hline
\textbf{Statistic} & \textbf{Value} \\
\hline
Number of topics & 60 \\
Total number of questions & 2,831 \\
Training questions & 1,819 \\
Development questions & 400 \\
Test questions & 612 \\
Time span & 2016-06 to 2025-11 \\
Unique event mentions & 913 \\
Timeline events & 865 \\
Average documents per topic & 19.7 \\
Average topic evidence length & 28,047 tokens \\
Average document length & 1,088.6 tokens \\
\hline
\end{tabular}
\caption{Core statistics of the AER benchmark. Agreement statistics will be added after the final annotation analysis is completed.}
\label{tab:main_stats}
\end{table}

\begin{figure}[t]
\centering
\includegraphics[width=\linewidth]{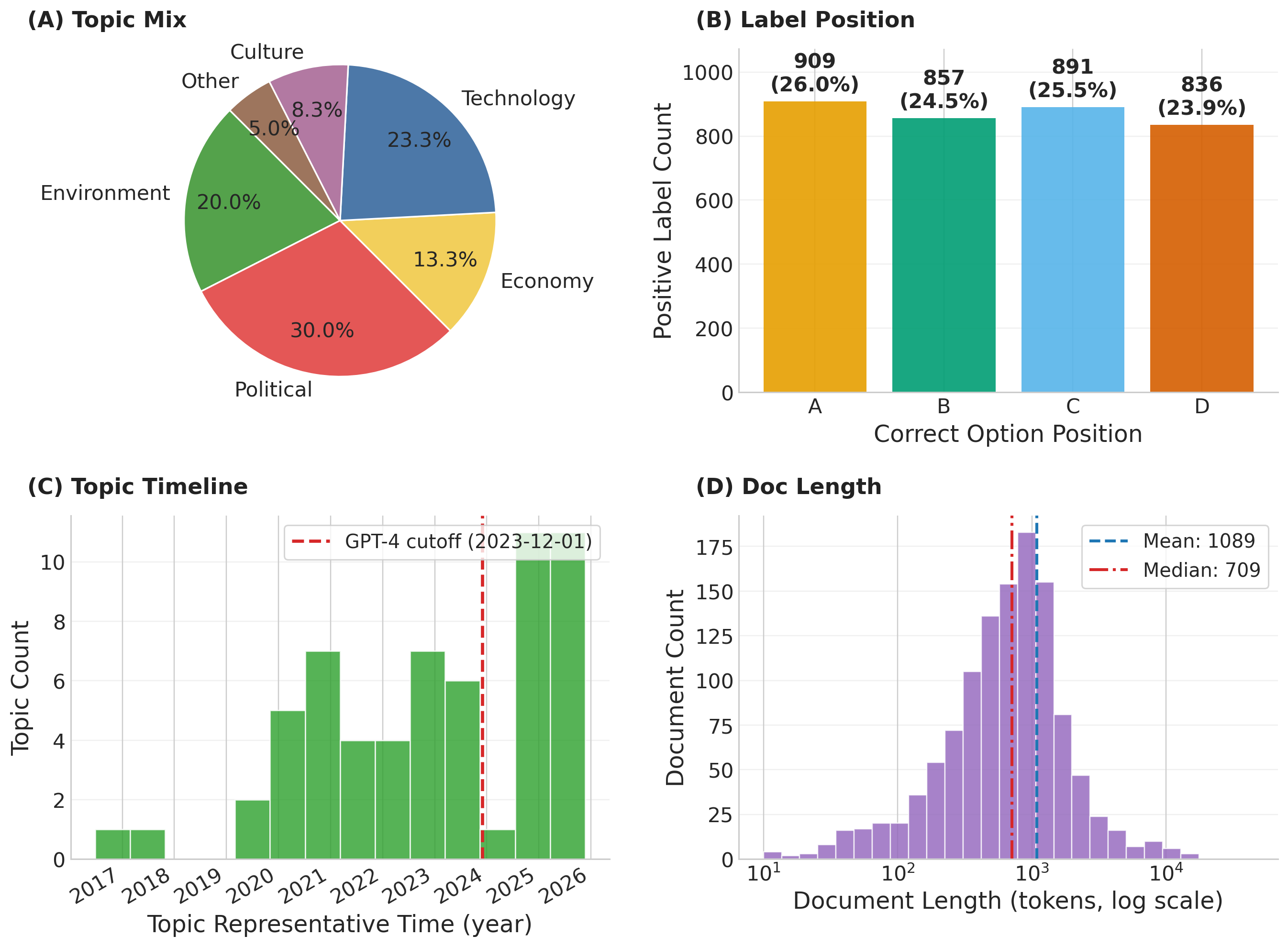}
\caption{Distributional overview of the  dataset: (A) topic composition across six categories, (B) frequency of correct labels at options A–D, (C) topic representative-time distribution with the GPT-4 knowledge cutoff (2023-12-01) marked by a dashed line, and (D) document length distribution in tokens, with mean and median indicated.}
\label{fig:data_stats}
\end{figure}

Table~\ref{tab:main_stats} summarizes the core statistics of the AER benchmark. Figure~\ref{fig:data_stats} further illustrates several key dataset distributions through four subfigures, including document category proportions, the positional distribution of correct answers, the temporal distribution of topic representation, and the document length distribution.

\subsection{Annotation Quality}

\label{sec:annotation_quality}

Constructing an evidence-based benchmark for direct cause identification is inherently challenging. In real-world events, causal relations are often not fully determinate. On the one hand, an outcome event may plausibly admit multiple candidate causes. On the other hand, the causal strengths of different candidates may be highly similar, making them difficult to distinguish consistently \citep{halpern2016actual}.

Moreover, topic-level document retrieval inevitably introduces some noise, and several stages of the pipeline---including event extraction, timeline construction, and candidate event selection---partly rely on LLM outputs. Human verification is therefore essential for ensuring dataset quality.

To improve annotation reliability, we asked three annotators to independently assign a three-way label to each candidate event pair. Specifically, given the available evidence, annotators labeled the relation between a candidate cause event and a target event as one of the following: 

(1) \textit{No causal relation}: the candidate event cannot be regarded as a direct cause of the target event.

(2) \textit{Moderate causal relation}: there is some direct causal connection between the two events, but the supporting evidence is limited in strength or exclusivity.

(3) \textit{Strong causal relation}: the candidate event is a clear and strongly supported direct cause of the target event.

We adopt Krippendorff's $\alpha$ as the primary inter-annotator agreement metric \citep{krippendorff2022content}. The overall agreement among the three annotators on the three-way classification task is $\alpha = 0.51$, indicating moderate agreement.

\section{Baselines and Pilot Analysis}

Before the formal evaluation phase, we conducted a pilot study to assess the difficulty of the benchmark and establish reference baselines. The purpose of this study was not to define additional shared-task tracks, but rather to understand how current strong language models perform under different evidence conditions.

\subsection{Baseline Models}

We evaluated three strong language models in a zero-shot prompting setting: GPT-4~\citep{openai2023gpt4}, Qwen-2.5-72B-Instruct~\citep{qwen2025qwen25}, and GLM-4~\citep{zeng2024chatglm}. These models were selected to cover both leading proprietary systems and open-weight models with strong general reasoning capabilities. All models were asked to predict answers directly from the provided evidence and candidate options, without task-specific fine-tuning.

\subsection{Experimental Setup}

We tested the models on a subset of 200 instances under two input conditions:

(1) {Original text (Ori\_T)}: the full set of retrieved documents, including distractor documents;

(2) {Summarized text (Sum\_T)}: timeline-style summaries generated by prompting Gemini 2.0 \citep{google2025gemini20flash}, which preserve the core event evolution while reducing redundancy and removing much irrelevant material.

These settings were used only for pilot analysis and baseline comparison, and do not constitute separate official tracks of the shared task.

\subsection{Pilot Results}

Table~\ref{tab:pilot} presents the pilot results. Across all tested models, performance drops when the input is changed from summarized evidence to the original document collections, indicating that evidence noise and long-context burden substantially affect direct cause inference.

\begin{table}[h]
\centering
\footnotesize
\begin{tabular}{lcc}
\toprule
\textbf{Model} & \textbf{Ori\_T} & \textbf{Sum\_T} \\
\midrule
GPT-4 & 68.66 & 70.35 \\
Qwen-2.5-72B-Instruct & 53.65 & 60.72 \\
GLM-4 & 58.12 & 60.36 \\
\bottomrule
\end{tabular}
\caption{Pilot results on a 200-instance sample under original-document input (Ori\_T) and summarized input (Sum\_T).}
\label{tab:pilot}
\end{table}

\subsection{Preliminary Findings}

The pilot study reveals two central difficulties captured by AER. First, systems struggle to locate and integrate causal evidence within long, noisy multi-document contexts. Even strong models perform noticeably better when the evidence is compressed into cleaner summaries. Second, systems are vulnerable to semantic confounds: when exposed to the full evidence collection, they are more easily misled by background conditions or related events that are semantically similar but not actually causal.

These findings support the motivation behind the benchmark design. The difficulty of AER does not arise merely from asking systems to select from a set of options; rather, the real challenge lies in the fact that relevant evidence is scattered, noisy, and mixed with plausible distractors.

\section{Participation and Results}

\subsection{Participation Overview}

The shared task was hosted on the Codabench platform and attracted broad participation throughout the evaluation period. In total, 122 participants registered for the task and made 518 submissions to the single official track. After the evaluation concluded, 21 teams submitted system description papers, among which 19 systems were retained as valid entries on the final leaderboard.

\subsection{Overview of Submitted Systems}

The submitted systems can be broadly categorized into four methodological families.

\paragraph{Retrieval-centered, evidence-grounded pipeline systems.}
These systems typically separate evidence selection from final decision making explicitly, often combining sparse retrieval, dense retrieval, reranking, and evidence verifiers or cross-encoder reasoning modules. Representative examples include the winning system {AILS-NTUA}, which uses graph-based retrieval followed by large language model reasoning and post-hoc correction through consistency constraints, and the {University of T\"ubingen} system, which combines option-oriented retrieval with an independent cross-encoder verifier. Similar retrieval--verification designs also appear in systems such as {uir-cis}, {sutta}, {thiyaga6851}, {AI4PC-Howard University}, and {X-NLP}. Overall, these systems share the intuition that the main bottleneck of AER lies in filtering out distractors and locating direct causal evidence before reasoning begins.

\paragraph{LLM prompting-based methods.}
These methods rely on large instruction-tuned models and often incorporate carefully designed prompts, role-based reasoning, self-consistency, or constrained decoding. For example, {AILS-NTUA} combines structured prompting, reflective prompt evolution, and self-consistency; {uir-cis} emphasizes high-precision decoding strategies to suppress over-selection. {Paradise}, {pushkar}, and {X-NLP} likewise treat prompt engineering, chain-of-thought reasoning, or retrieval-augmented prompting as core components. Such systems generally frame abductive reasoning as a controlled inference problem in which output calibration is at least as important as raw reasoning ability.

\paragraph{Supervised fine-tuning and distillation-driven systems.}
Unlike methods that mainly rely on inference-time prompting, these systems directly train discriminative models or compact student models on the task data. {CausalMinds} is the clearest example: it reports that simply fine-tuning GPT-4.1-mini with option-shuffling augmentation outperforms more complex prompting and verification pipelines. {HCMUS\_RepeatedGames} combines hybrid retrieval with extended LoRA fine-tuning \citep{hu2022lora} of a 32B model, while {YNU-HPCC} uses a DeBERTa-based\citep{he2020deberta} evidence-aware classifier together with teacher distillation. {KDW} also partially belongs to this category, as it employs a lightweight student model obtained through knowledge distillation. Taken together, these systems suggest that targeted supervision can sometimes compensate for architectural simplification on this benchmark.

\paragraph{Knowledge-enhanced and neuro-symbolic methods.}
These methods attempt to strengthen textual reasoning with explicit causal structure, such as evidence graphs, knowledge graphs, or theorem-proving-style verification processes. {KDW} combines knowledge-graph-based evidence extraction with distillation; {OseiBrefo-Liang} integrates hybrid retrieval, causal graphs, and neuro-symbolic policy optimization; and the {Newcastle} submission explores COMET-enhanced causal knowledge graphs together with Euclidean and hyperbolic embeddings. Although methodologically interesting, these approaches are, overall, less consistently competitive than the strongest retrieval-centered or fine-tuning-based systems.

Overall, the submitted systems exhibit a clear methodological pattern: most teams did not treat the task as ordinary multiple-choice question answering, but instead decomposed it into subproblems such as evidence selection, per-option verification, calibration, and structured set prediction.

\subsection{Leaderboard Results}

\begin{table}[t]
\centering
\small
\begin{tabular}{c l c}
\toprule
\textbf{Rank} & \textbf{Team Name} & \textbf{Score} \\
\midrule
1  & AILS-NTUA                & 0.95  \\
2  & d-itlab                  & 0.91   \\
3  & HCMUS\_RepeatedGames     & 0.90   \\
4  & KDW                      & 0.88   \\
5  & CausalMinds              & 0.88   \\
6  & X-NLP                    & 0.84   \\
7  & Younghee Jeong et al. & 0.84 \\
8  & uir-cis                  & 0.80  \\
9  & Paradise                 & 0.79   \\
10 & Bolun Liang et al.       & 0.78 \\
11 & sutta                    & 0.76 \\
12 & pushkar                  & 0.74   \\
13 & YNU-HPCC                 & 0.73   \\
14 & Clutch or Cry            & 0.72   \\
15 & OseiBrefo-Liang          & 0.61   \\
16 & Thiyaga6851              & 0.56 \\
17 & BBgame                   & 0.54 \\
18 & AI4PC-Howard University  & 0.53   \\
19 & REGLAT                   & 0.30    \\
\bottomrule
\end{tabular}
\caption{Final leaderboard of the shared task.}
\label{tab:leaderboard}
\end{table}

Table~\ref{tab:leaderboard} presents the final leaderboard. The best-performing system, {AILS-NTUA}, achieved a score of 0.95, establishing a clear lead over the rest of the field. The next two teams, {d-itlab} and {HCMUS\_RepeatedGames}, reached 0.91 and 0.90, respectively, showing that multiple methodological paths can achieve strong performance. Beyond this leading group, performance declines more gradually, forming a relatively dense middle tier roughly between 0.73 and 0.84, with lower-ranked systems forming a long tail.

This distribution supports two important observations. First, the task is largely solvable when systems can explicitly control distractors and define clear prediction boundaries. Second, the substantial gap between top and lower-ranked systems indicates that seemingly small design choices---especially in evidence selection, option isolation, and decoding calibration---can have major effects under the official evaluation metric.

The strongest systems exhibit distinct but convergent strategies. The top-ranked {AILS-NTUA} system combines graph-based retrieval, structured LLM reasoning, self-consistency, and iterative post-hoc consistency heuristics. This design directly addresses several failure modes in AER, including distractor documents, repeated options, and logically inconsistent multi-label outputs. {d-itlab} adopts a highly precision-oriented architecture in which candidate options are evaluated independently via multi-stage LLM gating and then combined with surprisal features in a final XGBoost ensemble \citep{chen2016xgboost}. This system is explicitly optimized to avoid false positives, which aligns closely with the task metric. In contrast, {HCMUS\_RepeatedGames} combines hybrid retrieval with extended LoRA fine-tuning of a 32B model, showing that large models can be highly effective when strong supervised adaptation is paired with evidence selection.

Among the remaining high-performing systems, {KDW} employs a logic-driven pipeline that combines knowledge-graph-based evidence extraction, distillation, and explicit handling of the {None of the others} option, whereas {CausalMinds} demonstrates that simple fine-tuning with option-shuffling augmentation can match or surpass more complicated multi-stage pipeline systems. Notably, {KDW} and {CausalMinds} achieved the same score as some more complex multi-stage systems. This suggests that success on this task does not depend on a single architecture, but rather on whether a system can reliably distinguish direct causes from semantically plausible distractors while avoiding over-prediction.

\subsection{Methodological Trends}

\paragraph{The dominance of retrieval and evidence compression.}
Many teams observed that the document packages provided by the task contain substantial irrelevant or only weakly relevant material. As a result, an increasing number of systems adopted hybrid retrieval, chunking, reranking, graph traversal, or option-oriented retrieval to reduce attention to distractors before final reasoning. This pattern is especially visible in top-ranked and upper-middle-ranked systems.

\paragraph{The importance of precision-oriented decision mechanisms.}
Because the evaluation setup awards no credit when incorrect options are included, many teams explicitly biased their systems toward more conservative prediction behavior. This tendency is reflected in threshold tuning, per-option verification, majority voting, constrained decoding, and post-hoc logical filtering. In other words, the evaluation setting encourages systems to behave more like cautious verifiers than aggressive generators.

\paragraph{The relative roles of prompting and fine-tuning.}
Prompt-based systems are widespread and remain competitive when combined with retrieval and calibration. However, the results also suggest that prompting alone is not always sufficient. Some teams reported that, in certain settings, chain-of-thought or multi-stage prompting could even hurt performance by amplifying spurious reasoning paths. By contrast, fine-tuned systems such as {CausalMinds} and {HCMUS\_RepeatedGames} show that direct supervision, even when combined only with relatively simple augmentation strategies, can yield highly competitive performance.

\paragraph{The promise and current limits of knowledge-enhanced reasoning.}
remains a promising but comparatively less mature direction. Several teams explored knowledge graphs, causal graphs, neuro-symbolic verification, or theorem-proving-style reasoning, but these approaches did not consistently outperform strong retrieval-plus-verification or fine-tuning baselines. This suggests that explicit structure is useful only when it materially improves evidence grounding; otherwise, it may simply introduce an additional layer of modeling complexity.

\subsection{Discussion}

The shared task suggests three main conclusions. First, AER is fundamentally constrained by evidence selection: systems must identify evidence for a \textit{direct} cause before effective reasoning is possible. Second, leaderboard performance reveals a trade-off between expressive generation and calibrated decision making; systems with explicit verification and conservative prediction often proved more reliable than unconstrained generative approaches. Third, the strongest submissions treated AER as a pipeline task, decomposing it into retrieval, evidence filtering, candidate verification, and constrained answer selection.

Overall, the results suggest that progress on AER depends less on model scale alone than on evidence grounding, robustness to distractors, and careful calibration under multi-label evaluation.

\section{Conclusion}

This paper introduced SemEval-2026 Task 12, Abductive Event Reasoning (AER), a shared task for direct-cause inference over noisy multi-document evidence. We presented the task formulation, dataset construction pipeline, evaluation protocol, pilot baseline analysis, and an overview of submitted systems and leaderboard results.

The shared task results show that strong performance depends not only on model capacity, but also on effective evidence selection, distractor filtering, and calibrated multi-label prediction. The strongest systems consistently treated AER as an evidence-grounded reasoning pipeline rather than as standard multiple-choice classification.

We hope AER will serve as a useful benchmark for future research on causal reasoning and multi-document understanding.





\section*{Acknowledgments}
This work is supported by the National Natural Science Foundation of China (No. U24A20335, No. 62406321). This work is also supported by Beijing Natural Science Foundation (L243006).

\bibliography{custom}

\appendix



\end{document}